\documentclass[conference,hidelinks]{ieeeconf}  % Comment this line out if you need a4paper

\IEEEoverridecommandlockouts                              % This command is only needed if 
\usepackage{graphicx} % for pdf, bitmapped graphics files
\usepackage{epsfig} % for postscript graphics files
\usepackage{amsmath} % assumes amsmath package installed
\graphicspath{{./Figures/}}
\usepackage{gensymb}
\usepackage{subcaption}
\usepackage{float}
\usepackage{breqn}
\usepackage{color}
\usepackage{wrapfig}
\usepackage{hyperref}
\usepackage{cite}
\usepackage{times}

\title{\LARGE \bf
Analysis of Rigid Extended Object Co-Manipulation by Human Dyads: Lateral Movement Characterization
}

\author{Erich A. Mielke \: Eric C. Townsend \: Marc D. Killpack % <-this % stops a space
}

\begin{document}

\maketitle
\thispagestyle{empty}
\pagestyle{empty}

%%%%%%%%%%%%%%%%%%%%%%%%%%%%%%%%%%%%%%%%%%%%%%%%%%%%%%%%%%%%%%%%%%%%%%%%%%%%%%%%
\begin{abstract}
During co-manipulation involving humans and robots, it is necessary to base robot controllers on human behaviors to achieve comfortable and coordinated movement between the human-robot dyad. In this paper, we describe an experiment between human-human dyads and we record the force and motion data as the leader-follower dyads moved in translation and rotation. The force/motion data was then analyzed for patterns found during lateral translation only. For extended objects, lateral translation and in-place rotation are ambiguous, but this paper determines a way to characterize lateral translation triggers for future use in human-robot interaction. The study has 4 main results. First, interaction forces are apparent and necessary for co-manipulation. Second, minimum-jerk trajectories are found in the lateral direction only for lateral movement. Third, the beginning of a lateral movement is characterized by distinct force triggers by the leader. Last, there are different metrics that can be attributed to determine which dyads moved most effectively in the lateral direction.
\end{abstract}

%%%%%%%%%%%%%%%%%%%%%%%%%%%%%%%%%%%%%%%%%%%%%%%%%%%%%%%%%%%%%%%%%%%%%%%%%%%%%%%%
\section{Introduction}
\begin{figure}[hbt]
  \centering
  \includegraphics[width=0.7\linewidth]{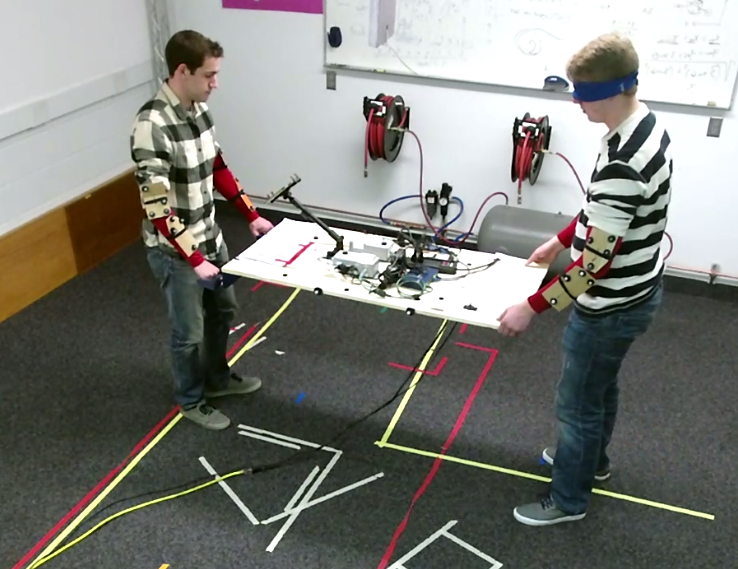}
  \caption{A dyad, one blind follower and one sighted leader, performs a co-manipulation task}
  \label{fig:example}
\end{figure}

Human-Robot Interaction (HRI) is an area of heavy interest in robotic studies. This is due to the combined strengths of a human-robot team: strength and execution from the robot and intelligence and planning from the human. Human teams are able to complete complex translational and rotational tasks, such as moving a table, couch, or other extended, rigid objects. These objects are heavy or unwieldy, and necessitate two or more people to carry them. A robot capable of replacing a human in these teams would help greatly in situations like search and rescue. Robots could help lift and remove rubble from disaster areas that would be impossible for human teams, and help take a victim on a stretcher to safety. Other applications include using robots to help load and unload moving vans, using robots to help move objects around warehouses, and many other co-manipulation applications where 2 person teams are used (see Fig. \ref{fig:example}.)

An important characteristic of these situations is a lack of definition. Often a task is poorly defined to one or both partners of a dyad, and a controller needs to be able to adapt to disturbances and trajectory changes. Thobbi et al. implemented a version of such a controller, although human intent was captured using motion capture, limiting its applicability \cite{Thobbi2011}. Intent is another issue of HRI co-manipulation. Many papers have suggested that haptic channels are an appropriate method of communication for human intent \cite{Basdogan2001},\cite{Noohi2016},\cite{Reed2007},\cite{Groten2013}. This makes sense, as human teams can move objects by interacting only through forces applied to the objects, rather than by communicating verbally or otherwise. Many studies have been done to conclude that robots can be controlled by human force input in this manner, but these studies involve upper-arm movements only, and often involve the human acting directly on the robot, and not through any extended object \cite{Corteville2007},\cite{Ikeura1997},\cite{Ikeura},\cite{Duchaine2007},\cite{TSUMUGIWA2002}.

While the studies and papers provided on co-manipulation prove that collaboration through force is applicable to some tasks, it is not clear that the algorithms and intent-estimators developed will work in less-defined scenarios. In order for a robot to work with humans, it needs to be able to respond in complex situations involving movement in 6 dimensions, 3 translational and 3 rotational. These tasks also involve whole-body motion and bi-manual manipulation by the participants. To understand how to design control methods for a robot in these tasks, one approach is to characterize the movements and forces produced by human-human dyads for a variety of tasks.

The main contributions of this paper are as follows
\begin{enumerate}
	\item Unique co-manipulation data from trials where human-human dyads moved a rigid table together (described in full in Section \ref{human_study})
	\item Observations on lateral movements from co-manipulation study, which include the following:
	\begin{itemize}
		\item Interaction forces are not minimized, but describe communication between partners
		\item Lateral trajectories resemble minimum-jerk trajectories, but only for lateral movements
		\item Evidence showing that lateral movements are triggered by a specific interaction force sequence
		\item Two possible metrics describing good lateral movement are minimizing angular velocity about superior axis and minimizing deviation from minimum-jerk trajectory
	\end{itemize}
\end{enumerate}

The paper is organized as follows. Section \ref{related work} explores relevant literature on human-robot co-manipulation tasks. The experiment described above is explained in full-depth in Section \ref{human_study}, including describing the equipment used, describing each task in detail, and describing the participants of the study. Section \ref{obsv} explores the main takeaways from the study. Last, Section \ref{conc} shows the conclusions of the paper and describes future work.

\section{Related Work}\label{related work}	
	Some researchers have suggested that haptic information is used to minimize a certain criterion. Flash and Hogan \cite{Flash1985}, for example, described human motion as following minimum-jerk trajectories for reaching movements, which has been used to describe motion objectives in many experiments \cite{Noohi2016}, \cite{Corteville2007}. The thought behind these experiments and theories is that for robots should move following minimum-jerk trajectories. A different criterion was suggested by Groten et al. \cite{Groten2013}, who focused on minimizing the energy of the motion. In this model, the robot always attempts to eliminate interaction forces, or forces that do not contribute to motion.
	
	Although the previously mentioned experiments show considerable promise for point-to-point, 1 DOF motion, many tasks require more DOF and less constrained motion. There are also tasks that do not fit well with the minimum-jerk trajectory \cite{Miossec2008}, \cite{Thobbi2011}. Humans often have no definite end goal when they are moving an object, only a general idea of which way the object should be travelling. In tasks such as these, a different approach is required. Ikeura et al. \cite{Ikeura2002}, \cite{Ikeura} developed a strategy for situations that required more flexibility in approach, such as tasks with no definite beginning or end. Their work sparked work in what is now known as variable-impedance control \cite{Ikeura1965}, \cite{Dimeas2015}. Other proposed models for HRI include programming by demonstration and finite state machines \cite{Gribovskaya2011}, \cite{Rozo2016}, \cite{Bussy2012a}, \cite{Bussy2012}.  

	Since physical HRI involves robots physically interacting with humans, studies have been performed with human subjects in hope of creating human-behavior based control for robots. Some studies involved shared virtual-environment loads \cite{Basdogan2001},\cite{Madan2015},\cite{Lawitzky2012}, others involved upper arm movements of individuals and dyads \cite{Wel2011},\cite{Reed2007}, and a few involved extended objects \cite{Karayiannidis2014}. These experiments clarify many aspects of physical HRI, including verifying that haptic information aids in co-manipulation tasks, noting some interaction patterns, and combining planning and learning to complete goal-oriented tasks. 
	
	Most of the related work has excellent performance for very specific tasks, but human motion is rarely 1-dimensional and usually involves coordination of body and arm motion. An approach to creating intuitive controllers for complex co-manipulation tasks is to study how human dyads perform such tasks, and a majority of previous studies have focused on simple tasks and objects without extent. We plan on developing an intuitive controller for physical HRI, and our approach is to focus on characterizing human dyad force and motion data through complex co-manipulation tasks. We developed an experiment to help us in our efforts. The purposes of this study were two-fold: first, to provide a baseline for how humans perform an 3-dimensional collaboration task on objects with extent, and second, to provide useful haptic information to use for creation of an intent estimator. Our study provided insights for collaborative motion of dyads not seen in other works, and formed a basis for developing a controller capable of handling complex tasks.
	
\section{Human Dyad Experiment}\label{human_study}
As a preliminary step in producing a 6-dimensional collaboration between a robot, we performed the following study involving human-human teams. If robots are to one day work alongside humans as partners, the robots need to perform tasks in a way that humans intuitively understand. Our approach is to base a HRI controller on concrete human characteristics in order to get performance that humans will understand and agree with.

\subsection{Experimental Setup}
After attaining IRB approval, we set up trials involving 2-person teams. These teams were to work together to perform a series of 6 object-manipulation tasks.
\subsubsection{Table}
The object the teams moved was a 59x122x2 cm wooden board -- meant to simulate an object (like a table) that is difficult for one person to maneuver. Attached to one end of the board were a pair of ABS 3D-printed handles, to which two ATI Mini45 force/torque sensors were fastened. The sensors transmitted data via ATI NET F/T Net Boxes, which passed data over ethernet to the computer at a rate of 100 Hz. 

\begin{figure}[hbt]
  \centering
  \includegraphics[width=1.0\linewidth]{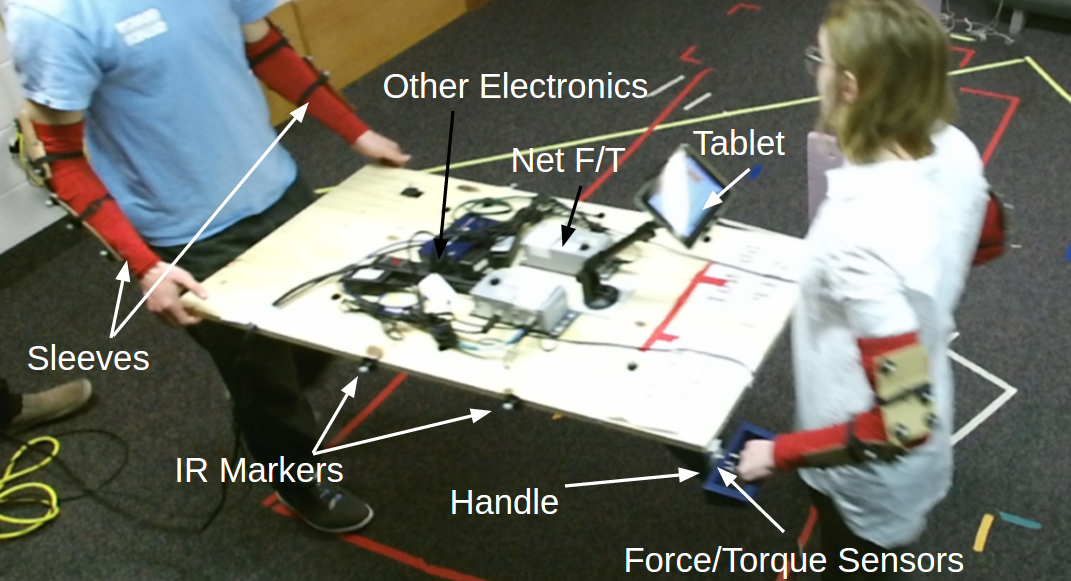}
  \caption{Setup for table and during trials}
  \label{fig:TableConfig}
\end{figure} 

The position of the board was tracked via Cortex Motion Capture software with a Motion Analysis Kestrel Digital Realtime System. A total of 8 Kestrel cameras were used to track 8 infrared markers placed on the board.  Using a static global frame established by the motion capture system, the position and orientation of the board could be tracked over time, and the force and torque data could be transformed into the board's frame as well as the static frame. The motion capture data was collected at a rate of 200 Hz.

Along with the infrared markers and force/torque sensors, the board also held an ethernet switch, a power strip, and all cables necessary for power and communication. One experimenter was tasked with making sure all no obstacles would trip the subjects, including moving these cables as necessary without exerting forces on the table. During the trials, a tablet was mounted on the board to display instructions to the participants. In total, the board weighed 10.3 kg. A visual of the board can be seen in Fig. \ref{fig:TableConfig}.
\subsubsection{Subjects}
The trial participants were outfitted with polyester arm sleeves for both arms. Two groups of four infrared markers were placed on rigid plates, and then attached to the sleeve, one on the upper arm and one on the lower arm. A blindfold was also used for the tasks where no communication was allowed.
\subsubsection{Arena}
\begin{figure}[hbt]
  \centering
  \includegraphics[width=0.7\linewidth]{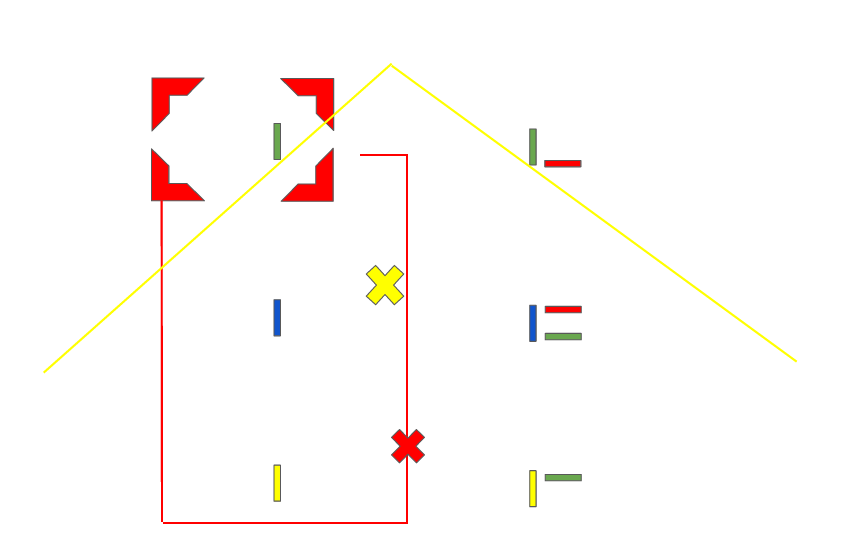}
  \caption{Colored tape for task delineation}
  \label{fig:TaskDelineation}
\end{figure}

The test arena was a volume measuring 490x510x250 cm. A series of colored tape lines (see Fig.~\ref{fig:TaskDelineation}) were placed on the floor of the volume, indicating key positions for each of the 6 object-manipulation tasks. On 3 of the walls surrounding the arena, we placed green, orange, and purple poster boards to help orient the leader when looking at the table. As seen in Fig. \ref{fig:AllStart} and Fig. \ref{fig:AllEnd}, there are colored bars on the edges of each task figure representing the walls with the corresponding color. This way, the leader could more easily determine the frame of reference of the table instructions.

The arena also was equipped with a video capturing device. The device we used was a Kinect 2, which allowed us to capture 3D point cloud data, as well as color video of each trial. Although we did not need the point cloud data for our purposes, we recorded extra data that may be useful in future work.

\subsection{Experimental Procedure}
\begin{figure}[hbt]
  \centering
  \fbox{\includegraphics[width=0.8\linewidth]{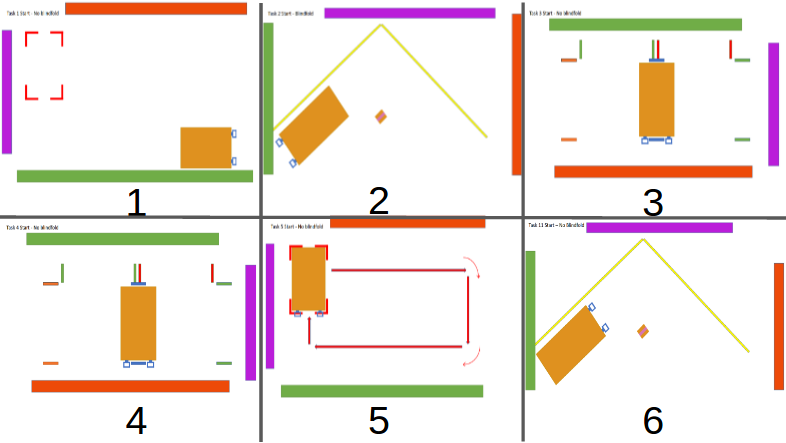}}
  \caption{Starting position for all tasks -- Tablet views}
  \label{fig:AllStart}
\end{figure}

\begin{figure}[hbt]
  \centering
  \fbox{\includegraphics[width=0.8\linewidth]{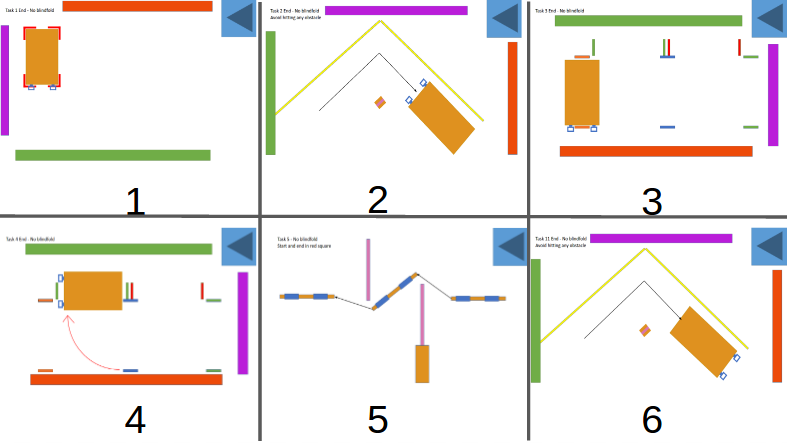}}
  \caption{Ending position for all tasks -- Tablet views}
  \label{fig:AllEnd}
\end{figure}

The experiment proceeded as follows. First, the participants were oriented on the purpose of the research and signed release forms. Second, a leader was chosen at random (by coin flip.) Third, each participant put on the sleeves and the participant designated as the follower placed the blindfold on their head, but not covering their eyes until they were about to perform a blindfolded task. Fourth, two preliminary test runs were performed by the participants with the researchers supervising. These test runs walked the participants through each motion required by the tests -- that is translation in x,y, and z axes and rotation in x,y, and z axes. The first run was done without the follower blindfolded, and the second was with the follower blindfolded. Fifth, the leader then was oriented on following the task instructions via the tablet on the table (see Fig. \ref{fig:AllStart} and Fig. \ref{fig:AllEnd}.) The researchers would display the task with visual instructions on the tablet, which corresponded to the colored tape on the ground. The leader then followed the instructions as outlined. Sixth, the 6 tasks were run 6 times. The tasks were split evenly between blindfolded and non-blindfolded, and were randomized in order for each group of participants. For instance, a group might perform task 1 non-blindfolded, followed by task 4 blindfolded, followed by 3 blindfolded, and so on. A researcher changed the setup between tasks, and two other researchers ran data collection for motion capture, force/torque, and video. Last, the participants were debriefed, filled out a questionnaire about the trials, and were paid.

The trials were designed in order to mimic standard motions that humans use when collaborating on moving an object. In order for a human to trust a robot to perform as expected, we attempted to establish a baseline behavior that described the actions of most participants. The tasks are outlined as follows and correspond to the numbers in Fig. \ref{fig:AllStart} and Fig. \ref{fig:AllEnd}: 
\begin{enumerate}
\item Pick and Place
\begin{itemize}
\item Translation and rotation, but emphasizing the placement of the object, like placing an object in a specific location and orientation
\end{itemize}
\item Rotation and Translation -- Leader facing backwards
\begin{itemize}
\item Rotating board in one axis while translating object, like moving an object around a corner in a hallway with the leader walking backwards
\end{itemize}
\item Pure Translation
\begin{itemize}
\item Translation in one axis, like both partners moving laterally with an object
\end{itemize}
\item Pure Rotation
\begin{itemize}
\item Rotation in one axis, like one parter rotating around other
\end{itemize}
\item 3D Complex Task -- Translation and Rotation in multiple axes
\begin{itemize}
\item Moving object with translation in all three axes while avoiding certain 3D obstacles, like moving an object through complex spaces
\end{itemize}
\item Rotation and Translation -- Leader facing forwards
\begin{itemize}
\item Rotating board in one axis while translating object, like moving an object around a corner in a hallway with the leader at the front
\end{itemize}
\end{enumerate}

The physical carrying-out of the task started with each participant grasping an end of the board, the leader by the end with sensors and the follower by the end without sensors. They would then lift the table and the follower try to follow the leader as the leader performed the task indicated on the tablet. Once they reached the position, they set the board back on the ground and released. This constituted a single trial.% A sample of task 5 being performed blindfolded and not blindfolded can be seen at \url{https://youtu.be/DAbLRDN20yE}. This task is shown as it encapsulates a majority of the motions seen in all the tasks as they were performed.
\subsection{Data Collection}
A total of 21 trials were performed, and participants for the trials were recruited using flyers, social media, and word-of-mouth. Trials occurred during February and March of 2016. The participants were comprised of 26 men and 16 women of ages 18-38, and the average age was 22. There were 38 right-handed and 4 left-handed. A scheduling website was used to facilitate trial sessions, and participants signed up for an available slot on the site.

If, during a task, any error occurred -- such as participants performing a task incorrectly or a fault in data collection -- the task would be stopped and repeated.
\subsection{Data Analysis}
\begin{figure}[hbt]
  \centering
  \includegraphics[width=0.8\linewidth]{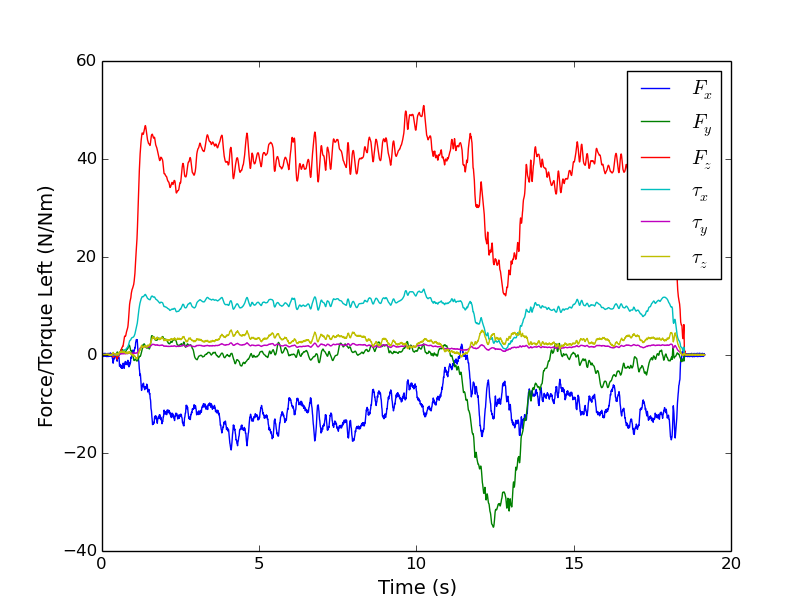}
  \caption{Raw data from force/torque sensor 1}
  \label{fig:f1raw}
\end{figure}

\begin{figure}[hbt]
  \centering
  \includegraphics[width=0.8\linewidth]{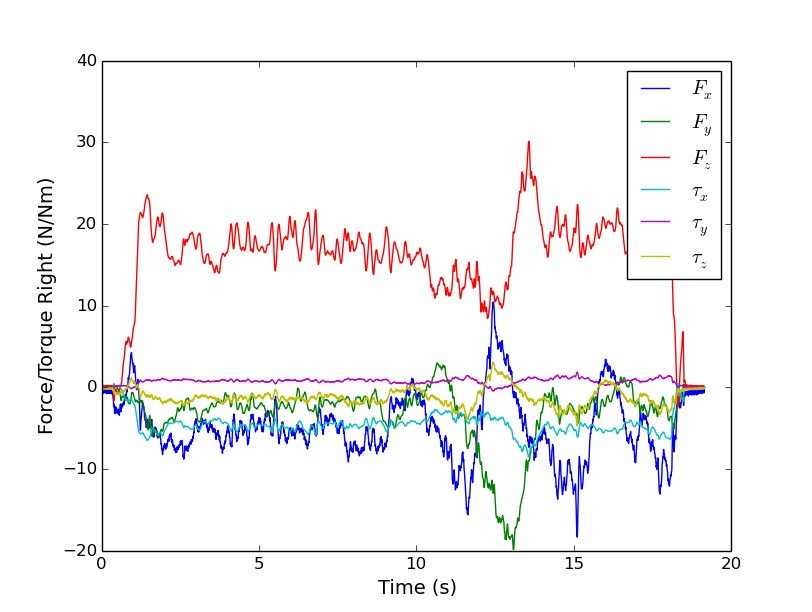}
  \caption{Raw data from force/torque sensor 2}
  \label{fig:f2raw}
\end{figure}

\begin{figure}[hbt]
  \centering
  \includegraphics[width=0.8\linewidth]{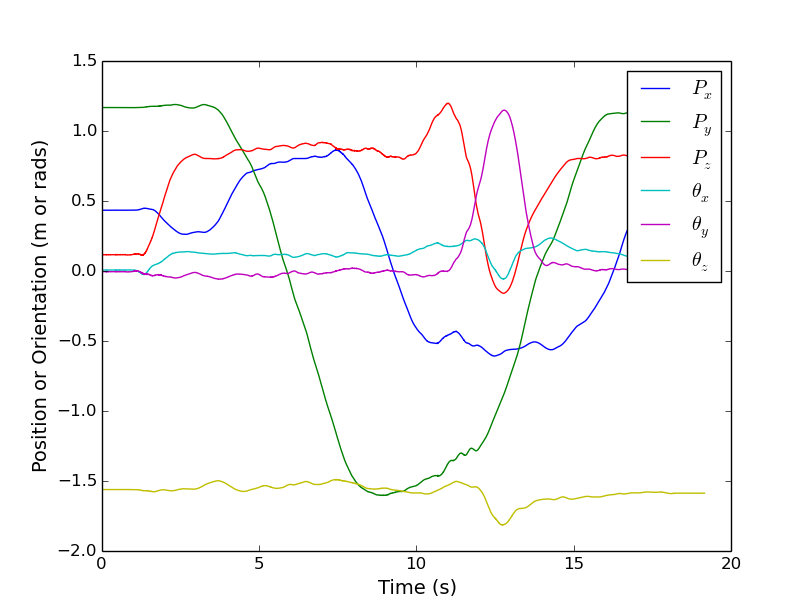}
  \caption{Raw data of table pose from motion capture}
  \label{fig:paraw}
\end{figure}

As previously stated, the data we acquired for each trial was the force and torque data from the two sensors on the handle, the position and orientation of the table, the position and orientation of the participant's arms, as well as the point cloud data from the Kinect 2. The data we were most interested in was the force and torque data in relation to the position and orientation of the table. A sample of the data collected for the 3D complex task can be seen in Figs. \ref{fig:f1raw} - \ref{fig:paraw}.

\section{Observations}\label{obsv}
\begin{figure}[hbt]
  \centering
  \includegraphics[width=0.8\linewidth]{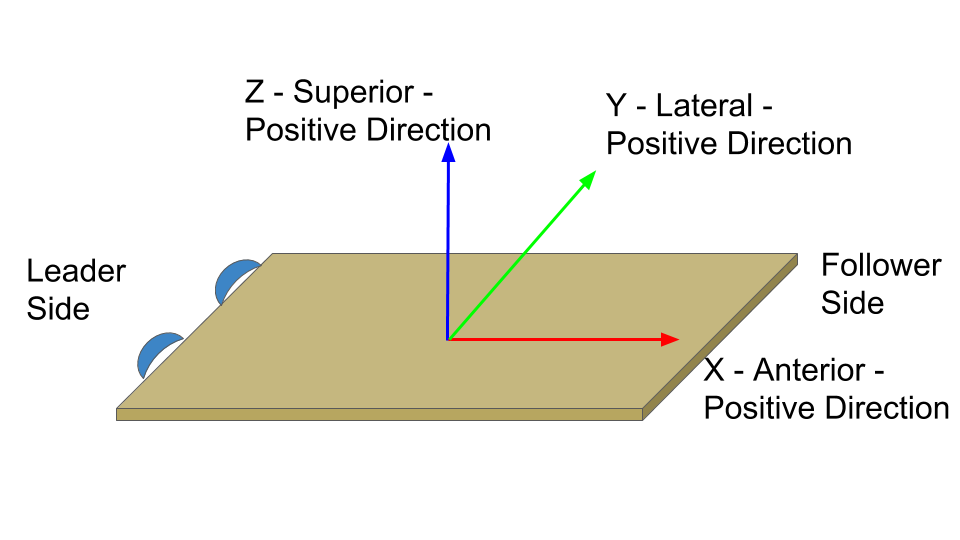}
  \caption{Anatomical direction reference with corresponding table axis. X is anterior, Y is Lateral, and Z is Superior.}
  \label{fig:tabdir}
\end{figure}

After completing the experiment, we focused on discovering patterns in the data. Although the experiment involved 6 different tasks, this paper targets characterizing the blind versions of task 3 -- Pure Translation. The emphasis was placed on this task for a couple reasons. First, as discussed in Section \ref{related work}, most research done in this area of HRI involved either lateral movement with no extended object, or only anterior direction movements (see Fig. \ref{fig:tabdir} for directions reference.) When co-manipulating an extended object, there is ambiguity between lateral and rotational movements. Therefore, characterizing how humans are able to recognize a lateral movement with an extended object and distinguish it from other movements is key. Second, since other tasks involve this motion, knowing the defining characteristics of this motion helps to recognize it in more complex tasks. We are interested in only the blind tasks, since the haptic-channel communication method is the main condition that would be present in a human-robot dyad. 

\subsection{Interaction Forces}
As suggested by Noohi et al. \cite{Noohi2016}, interaction forces could be used as a source of communication. Interaction forces are the forces that do not directly relate to motion, i.e. the forces applied by each participant in opposing directions. In our study, the force/torque sensors could not discern between external forces -- forces that move the object -- and interaction forces, but rather measured the total force applied, so we calculated the interaction force after the experiment ended.

\begin{figure}[hbt]
  \centering
  \includegraphics[width=0.8\linewidth]{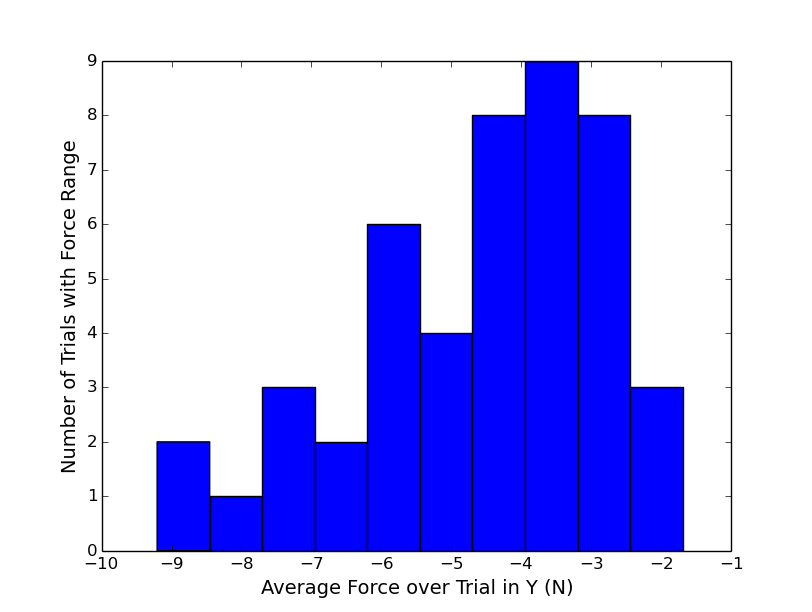}
  \caption{Histogram of lateral movement average force for trials}
  \label{fig:yhisto}
\end{figure}

\begin{figure}[hbt]
  \centering
  \includegraphics[width=0.8\linewidth]{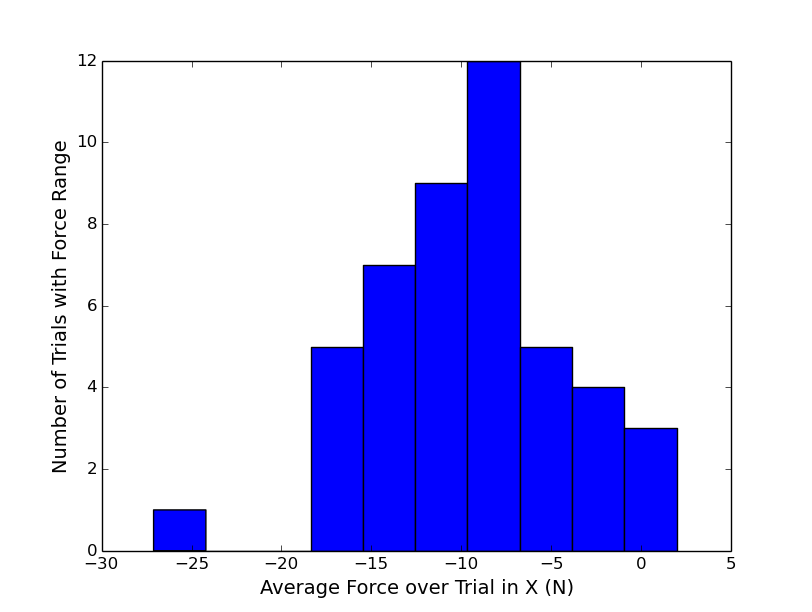}
  \caption{Histogram of anterior movement average force for trials}
  \label{fig:xhisto}
\end{figure}

Eq. \ref{ftot} shows how the total force was split up. $F_t$ is the total force on the object, $F_i$ is the interaction force, and $F_e$ is the external force. The motion capture data described the pose of the table over time, and was differentiated twice to acquire the acceleration data. With a known mass and acceleration, the external force was calculated (Eq. \ref{fext}), and removed from the total force to give us the interaction force for the task. 

\begin{equation}\label{ftot}
F_t = F_i + F_e
\end{equation}

\begin{equation}\label{fext}
F_e = ma
\end{equation}

For the anterior, $X$, and lateral, $Y$, directions, the only external force being applied is the force applied from the participants, whereas in the vertical $Z$ direction, gravity was also applied. For all calculations and analysis in this paper, the forces are filtered to 10 Hz to represent human response ranges. The muscle response of humans can reach up to 100 Hz for brief, forceful efforts, but often falls in the range of 1-10 Hz \cite{Burdet}.

As mentioned in Section \ref{related work}, some prior work in HRI has had the goal of minimizing interaction forces \cite{Groten2013}. Our study showed that this may not always be the case. For lateral movements, we calculated the average interaction force in the anterior and lateral directions. Histograms showing the distribution across all trials of the average interaction force of an individual trial are shown in Fig. \ref{fig:yhisto} and \ref{fig:xhisto}. As can be seen, the interaction force was almost always non-zero for a lateral movement, in both the anterior and lateral directions. In the lateral direction, the interaction force was applied toward the direction of desired motion, indicating the participants were resisting lateral motion. Also interesting, the anterior interaction force was much higher in magnitude than the lateral interaction force. The force was also in the negative direction, indicating that the participants are pulling the object away from each other as they move laterally (see Fig. \ref{fig:tabdir} for clarity on directions.) It is not clear why this average force was so substantial in the anterior direction, but here are our hypotheses:

\begin{enumerate}
	\item These forces were used for object and human stability
	\item These forces were used to communicate intent
\end{enumerate}

We will conduct more research with respect to the stability hypothesis, but we will discuss the intent hypothesis further in Section \ref{lmsc}. This result is significant because it implies that lateral collaborative movements do not rely only on forces applied in the lateral direction, but also on forces in the anterior direction, which is not seen in many state-of-the-art HRI controllers. Another takeaway is that minimizing interaction forces may not yield results easily understood by human partners in co-manipulation tasks, since it is now evident that humans are not minimizing these forces, at least not in all directions.

\subsection{Minimum-Jerk}
\begin{figure}[hbt]
  \centering
  \includegraphics[width=0.8\linewidth]{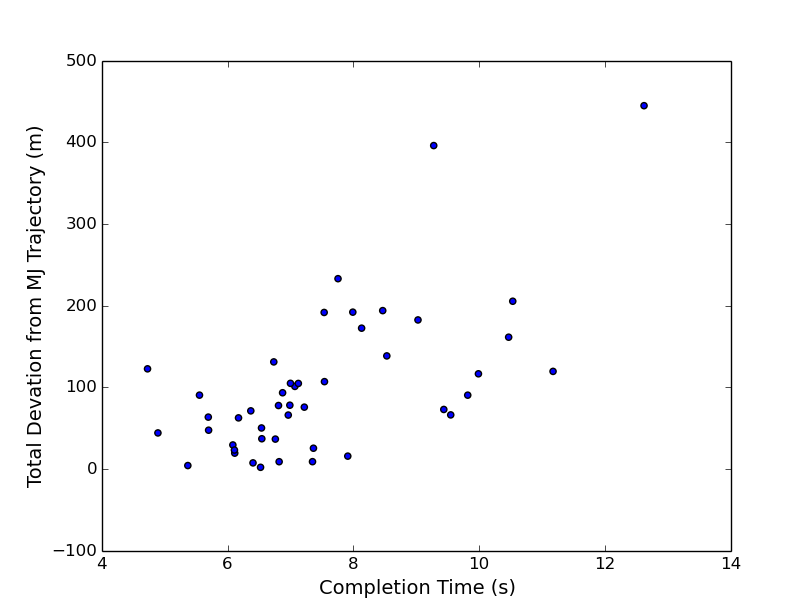}
  \caption{Comparing completion time to deviation from MJ trajectory}
  \label{fig:metcom}
\end{figure}

The minimum-jerk (MJ) movement has been well-documented to be a basis for human arm movements, especially in point-to-point movements. We did not expect this movement would appear for these tasks, since one participant was blindfolded and unaware of the task specifications, and also were using whole-body motion rather than arm-only motion. However, another interesting finding from our study was that the lateral movement tasks resembled a MJ movement in the lateral direction, especially for the dyads that completed the task more quickly. Fig. \ref{fig:metcom} shows the correlation between deviation from MJ trajectories and an increase in time to complete the task. The slower tasks often had a larger error between their position, and the ideal MJ position, whereas the quicker tasks generally had a smaller error.

\begin{figure}[hbt]
  \centering
  \includegraphics[width=0.8\linewidth]{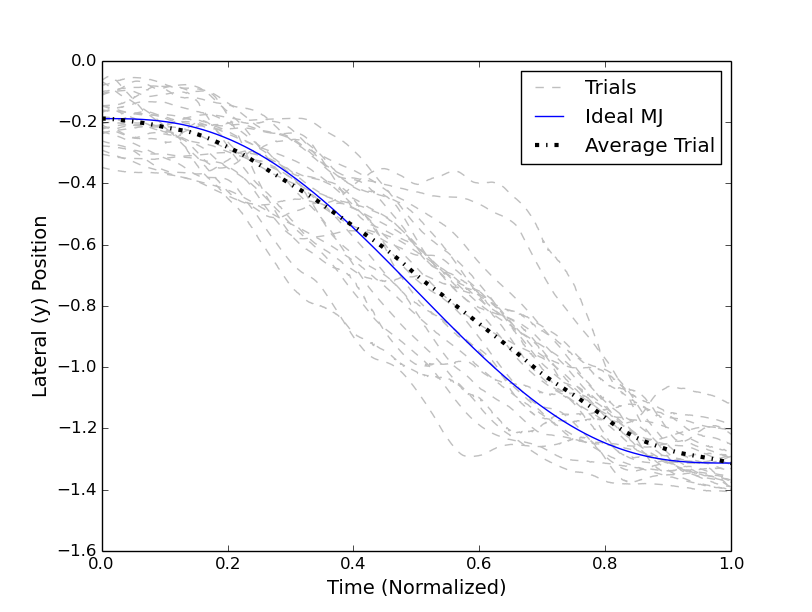}
  \caption{Lateral trial trajectory with ideal MJ and average trajectories}
  \label{fig:negmj}
\end{figure}

\begin{figure}[hbt]
  \centering
  \includegraphics[width=0.8\linewidth]{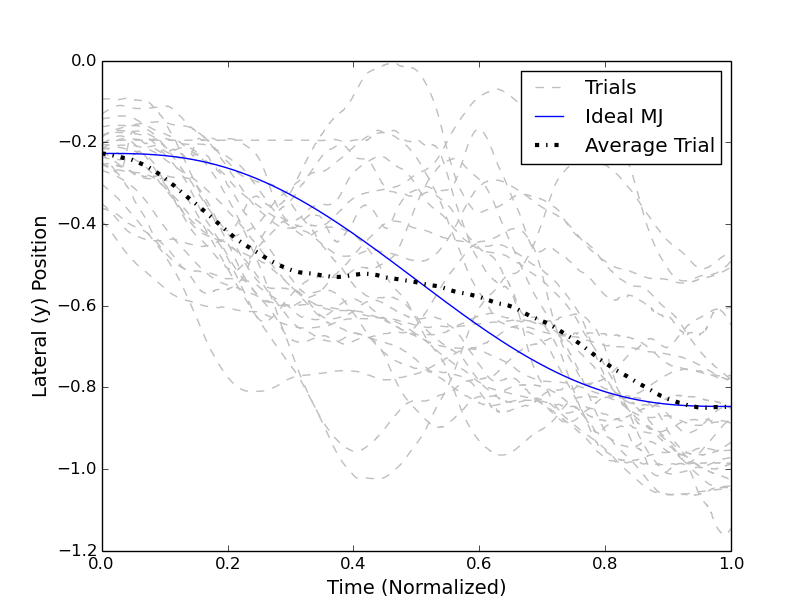}
  \caption{Rotational trial trajectory with ideal MJ and average trajectories}
  \label{fig:neg9}
\end{figure}

Overall, the lateral movement stayed close to the MJ trajectory, and adhering to a similar trajectory over all trials corroborates the results of similar 1-dimensional study \cite{Bussy2012a}. Fig. \ref{fig:negmj} shows the lateral position of the lateral tasks over time. The gray dotted lines show each individual task, the black dotted line is the average of all the tasks, and the blue line is the ideal MJ trajectory given an average start and stop position. As we can see, even though the follower did not know the end position, they stayed fairly close to the MJ trajectory in the lateral direction.

However, during the lateral tasks, there was also movement in the anterior direction, which did not follow a MJ trajectory. Additionally, there were tasks that did not adhere closely to a MJ trajectory due to disturbances. Comparing the lateral task with the blind rotational task (Task 4), see Fig. \ref{fig:neg9}, shows that achieving MJ trajectories may not always have been the goal of the dyads during coordinated motion. As we see, the average trajectory in Fig. \ref{fig:neg9} was significantly distinct from the MJ trajectory, and looking at each individual trial shows that there were a variety of dissimilar paths taken in the rotation. Thobbi et al. suggest, however, that using MJ as a basis for a controller is not ideal, as it is too restrictive \cite{Thobbi2011}. Our conclusion to this point is that MJ trajectories are useful in describing metrics (discussed more in Section \ref{metrics}.) MJ trajectories also seem to describe lateral movements fairly well, but not accurately describe the position of extended objects during rotational tasks.

\subsection{Lateral Movement Start Characteristics}\label{lmsc}

In the case of lateral movements, we recognized some patterns in how people behaved. Studying the videos of the lateral motion task, we saw that the follower often guessed the leader's intent wrong, and began to rotate when the leader started their movement. When this happened, the leader would flex their arm on one side of the table, causing a torque on the table, and the follower would then commence moving in the correct manner. Upon seeing this, we started to look for a pattern of applied torques that would indicate the start of a lateral movement. 

\begin{figure}[hbt]
  \centering
  \includegraphics[width=0.8\linewidth]{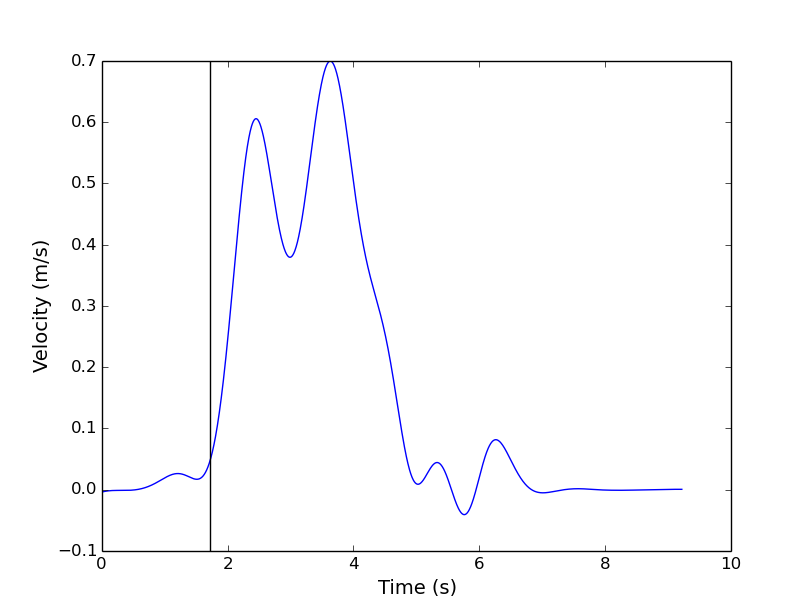}
  \caption{Velocity (lateral direction) with line delineating start point}
  \label{fig:righttrig}
\end{figure}

\begin{figure}[hbt]
  \centering
  \includegraphics[width=0.8\linewidth]{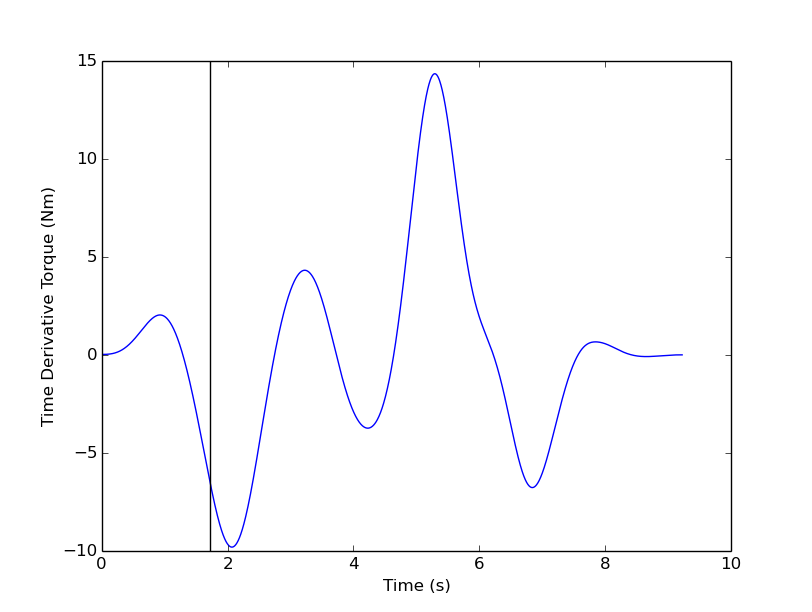}
  \caption{Trigger pattern (anterior direction) for time derivative of torque about $Z$ axis with line delineating start point}
  \label{fig:torquetrig}
\end{figure}

Fig. \ref{fig:torquetrig} illustrates the pattern we found. The leader applied forces, increasing on one hand and decreasing on the other hand, causing an increase in interaction torque about the $Z$ axis. The increasing magnitude of the slope, or time derivative of torque, signalled to the follower that the leader wanted to move laterally. After reaching a certain height (to indicate object was off the ground) and torque threshold,the follower moved and the desired lateral movement began. We then searched through each trial for the first instance of meeting the height and torque thresholds and noted the trigger time. We then determined whether the pattern held based on whether the table's $Y$ velocity at the trigger time matched the first instance of movement in the lateral direction for the trial, as shown in Fig. \ref{fig:righttrig}. This method was capable of correctly predicting the start time for 75\% (35 of 46) of all blind, task 3 trials with useable data. Also, this same method was not effective in predicting the start time for the rotational task, showing there is a distinction in how these tasks are triggered. It is not clear how the velocity is affected by the forces after this point, but this is an important open research question that we are still exploring in order to make a robot effective for co-manipulation.

Characterizing the force pattern causing a lateral movement is an important development in human-robot co-manipulation. Distinguishing between rotation and translation was difficult even for some of the participants of our trials. By determining what humans do to cause these movements, we can develop methods of communicating to the robot human intent more accurately than with previous methods. 

\subsection{Metric Observations}\label{metrics}
Our main objective is to identify models and triggers that will allow us to develop control algorithms for robots to successfully co-manipulate objects with extent. However, we need metrics of what performance behavior specifically the robots should be imitating. In order to quantify the quality or performance of an individual trial, we needed to determine metrics that were representative of the performance. This was a difficult exercise, since all subjects were able to complete the trials, and were given no specific directions other than for the leader to complete the on-screen objectives, and the follower to follow the leader. Possible metrics considered were:
\begin{itemize}
	\item Completion Time
	\item Distance Covered
	\item Average/Max Force
	\item Average/Max Velocity
	\item Average/Max Power
	\item Average/Max Angular Velocity
	\item Deviation from MJ Trajectory
\end{itemize}
These values were compared using the Pearson correlation coefficient, and surprisingly offered very little correlation. For the lateral task, the metrics that were most applicable were average/max angular velocity and deviation from the MJ trajectory. We expected a good task to be one where the table minimized the average angular velocity about the $z$ axis, and stayed relatively close to the MJ trajectory, but the correlation coefficient between these two metrics was 0.05. In fact, there was a much stronger correlation between deviation from MJ trajectory and completion time, average lateral velocity, and distance covered -- being 0.63,-0.42, and 0.38 respectively. Intuitively, minimizing angular velocity would be an ideal metric for this task, but the fact that deviation from MJ trajectory corresponds so well with the other metrics means it should be a point of emphasis in future research.

\section{Conclusion and Future Work}\label{conc}
In this paper, we have discussed the problems faced by many co-manipulation HRI controllers currently employed. We discussed the advantages of creating control methods based on human dyads to increase the ability of human-robot dyads to adapt to less-defined situations and described our experiment gathering the force and motion data for several simple and complex tasks involving human dyads. The main takeaways from this data are that interaction forces play an important role in communicating intent between dyads in co-manipulation and that they are likely not minimized as previously supposed. Lateral movements display characteristics of minimum-jerk movements in the lateral direction. These movements are triggered by a predictable pattern of forces, and there are multiple metrics that are important in considering what characterizes a good lateral movement. 

Our planned future work will include determining the pattern of the rotational movement start trigger, determining the stop trigger for rotational and lateral movements. After defining rotational versus lateral, a controller will be implemented on a robot platform and more experiments will be conducted to help confirm that humans are able to co-manipulate an object comfortably with a robot partner. 
%
%\section*{Acknowledgements}
%We would like to acknowledge and thank 
%%%%%%%%%%%%%%%%%%%%%%%%%%%%%%%%%%%%%%%%%%%%%%%%%%%%%%%%%%%%%%%%%%%%%%%%%%%%%%%%

\bibliographystyle{IEEEtran}
\bibliography{references}

\end{document}